\documentclass[conference]{IEEEtran}
\IEEEoverridecommandlockouts
\usepackage[T1]{fontenc}    
\usepackage{cite}
\usepackage{graphicx}
\usepackage{amsmath}
\usepackage{amssymb}
\usepackage{booktabs}
\usepackage{subcaption}

\usepackage{tabularx}
\usepackage{adjustbox}
	
\usepackage{color, colortbl}
\usepackage[normalem]{ulem}

\usepackage[natural]{xcolor}
\usepackage[markup=underlined]{changes}
\usepackage{todonotes}
\usepackage{mathtools}
\usepackage{url}
\usepackage{algorithm}
\usepackage{algorithmicx}
\usepackage[noend]{algpseudocode}
\usepackage{gensymb}
\usepackage{siunitx} 
\usepackage{interval}
\usepackage[pagebackref,breaklinks,colorlinks]{hyperref}
\usepackage{svg}
\usepackage{mathrsfs}
\usepackage{dsfont}
\usepackage{amssymb}
\usepackage{pifont}
\usepackage{stfloats} 
\usepackage{pdflscape}
\usepackage{afterpage}
\usepackage{multirow}

\title{LiDAR Based Semantic Perception for Forklifts in Outdoor Environments}

\author{Benjamin Serfling\textsuperscript{*}, Hannes Reichert\textsuperscript{*}, Lorenzo Bayerlein, Konrad Doll and Kati Radkhah-Lens 
	\thanks{B. Serfling, H. Reichert, K. Doll and K. Radkhah-Lens are with the Faculty of Engineering and Informatics,
		University of Applied Sciences Aschaffenburg, Aschaffenburg, Germany
		{\tt\footnotesize firstname.lastname@th-ab.de}}
	\thanks{L. Bayerlein is with the Linde Material Handling GmbH, Kion Group,
		Aschaffenburg, Germany
		{\tt\footnotesize lorenzo.bayerlein@kiongroup.com}}
    \thanks{\textsuperscript{*}The authors contributed equally.}
}

\begin{document}

\maketitle
 
\begin{abstract}

In this study, we present a novel LiDAR-based semantic segmentation framework tailored for autonomous forklifts operating in complex outdoor environments. Central to our approach is the integration of a dual LiDAR system, which combines forward-facing and downward-angled LiDAR sensors to enable comprehensive scene understanding, specifically tailored for industrial material handling tasks. The dual configuration improves the detection and segmentation of dynamic and static obstacles with high spatial precision. Using high-resolution 3D point clouds captured from two sensors, our method employs a lightweight yet robust approach that segments the point clouds into safety-critical instance classes such as pedestrians, vehicles, and forklifts, as well as environmental classes such as driveable ground, lanes, and buildings. Experimental validation demonstrates that our approach achieves high segmentation accuracy while satisfying strict runtime requirements, establishing its viability for safety-aware, fully autonomous forklift navigation in dynamic warehouse and yard environments.

\end{abstract}
\section{\large Introduction}
\label{sec_introduction}

\subsection{Motivation} \label{sec_motivation}
Autonomous guided vehicles (AGV) play an increasingly vital role in modern industrial operations. Although compact indoor vehicles, such as pallet lifters and platform AGVs, have achieved a high level of autonomy, larger vehicles, such as counterweight forklifts, still rely heavily on human operators. These forklifts are often required to navigate complex environments that span both indoor and outdoor settings, making full automation considerably more challenging.

One of the key technical hurdles in enabling autonomous forklift operation is semantic perception, including the accurate detection of drivable free space. This capability is fundamental to core tasks such as path planning, task allocation, and collision avoidance. Inaccurate semantic perception can lead to inefficient navigation, increased operational time, and potential safety hazards. It also impacts the vehicle's ability to detect and track obstacles like pedestrians, other vehicles, or dynamic objects. Achieving robust semantic perception in real-world warehouse environments is particularly challenging due to their unstructured nature, characterized by irregular layouts, movable obstacles such as pallets and racks, and the absence of clear lane markings. In addition, forklifts frequently transition between indoor and outdoor areas, where lighting and surface conditions can vary dramatically.

Moreover, automotive-grade sensors, that are designed for narrow fields of view and long-range detection on highways, are not suitable for forklift applications, which require up to 270 ° coverage and accurate short-range detection of objects approaching from the side, especially within tight warehouse aisles. Because no standard solution meets these needs, we propose a custom dual-LiDAR configuration and perception pipeline specifically optimized for the spatial and operational constraints of industrial logistics environments.

The use of rear wheel steering introduces further complications, as high angular velocity can produce motion blur in sensor data, reducing the effectiveness of vision-based methods.

LiDAR is a key technology for autonomous forklift, providing precise 3D perception that remains robust under varying lighting conditions, in contrast to camera-based systems. It enables the accurate detection of free space and obstacles in mixed indoor-outdoor unstructured environments. Recent industry trends indicate increasing LiDAR resolution and decreasing sensor costs, making high-performance LiDAR more accessible and practical for industrial AGV applications. This combination of precision, reliability, and affordability establishes LiDAR as a key enabler for safe and efficient autonomous navigation. However, the deployment of high-resolution LiDAR in real-world warehouse environments remains largely unexplored.

\subsection{Requirements} \label{sec_requirements}
\subsubsection{Real-Time Constraints}
Real-time processing is critical for autonomous forklifts operating in dynamic environments like warehouses. With two Ouster OS0-128 LiDAR sensors running at 10 Hz, each provides a full scan every 100 ms. We target a 30 Hz control loop for perception, allowing 33.3 ms to be allocated to process sensor data. This leaves additional runtime for downstream tasks, such as path planning and decision making.
\subsubsection{Segmentation Accuracy}
High segmentation accuracy is critical for safe and reliable autonomous forklift operation. Consistent identification of drivable ground is required for effective path planning, while precise detection of safety-critical classes, especially pedestrians, is essential for collision avoidance and emergency braking. Segmentation performance directly influences the system’s ability to make safe navigation decisions and respond to dynamic changes in the environment.
\subsection{Related Work} \label{sec_sota}
\subsubsection{Automotive LiDAR Semantic Segmentation}
Large-scale annotated datasets such as SemanticKITTI~\cite{behley2019iccv}, SemanticPOSS~\cite{pan2020semanticposs}, SemanticUSL~\cite{jiang2020lidarnet}, SemanticTHAB~\cite{reichert_2025_14906179}, nuScenes~\cite{fong2021panoptic}, and the Waymo Open Dataset~\cite{Waymo} have significantly advanced LiDAR-based semantic segmentation for autonomous driving. These benchmarks have fostered a variety of approaches aimed at optimizing the trade-off between segmentation accuracy and computational efficiency.

Spherical projection-based methods, such as FRNet~\cite{FRNet}, FIDNet~\cite{Zhao2021FIDNetLP}, and CENet~\cite{cheng2022cenet}, convert 3D point clouds into 2D range images, allowing fast inference through convolutional neural networks (CNNs) while preserving competitive mean Intersection over Union (mIoU) scores. More recent Vision Transformer (ViT)-based architectures, including RangeFormer~\cite{RangeFormer} and RangeViT~\cite{RangeViT}, further improve accuracy but at the cost of increased latency and reduced frame rates.

Alternatively, volumetric and raw point cloud-based methods such as SphereFormer~\cite{SphereFormer}, Cylinder3D~\cite{Cylinder3D}, and MinkUNet~\cite{tang2020searching} generally achieve higher segmentation accuracy, but are often unsuitable for real-time deployment due to computational complexity. The current state-of-the-art semantic segmentation accuracy on SemanticKITTI is held by the Point Transformer V3 foundation model~\cite{wu2024ptv3}. However, as highlighted by Reichert et al.~\cite{reichert2025realtimesemanticsegmentation}, many of these methods struggle to scale to high-resolution LiDAR sensors under strict real-time constraints. Their work introduces a revised spherical projection strategy coupled with a lightweight CNN backbone that enables real-time semantic segmentation of high-resolution scans.

\subsubsection{LiDAR-Based Perception in Material Handling}

Material handling perception tasks present a different set of challenges compared to autonomous driving, including unstructured environments, dynamic obstacles, and the need for rapid inference to support agile robotic behavior.

The SeMantic InDustry dataset~\cite{wang2024sfpnet} targets specifically the industrial domain. However, it is limited to static scene elements and does not include dynamic object classes such as forklifts or persons. Furthermore, the LiDAR sensors used have a low resolution compared to modern standards. The Goose-Ex dataset \cite{goose-ex-dataset} provides high-resolution annotated LiDAR scans for the operation of an excavator in unstructured outdoor environments.

Schreck et al.~\cite{surface_normals} addresses the challenges of material handling with a heuristic method for free space detection, tailored to forklifts operating in mixed indoor-outdoor spaces. Their approach uses height-change cues and surface normals computed from spherical projections to segment traversable regions in real time, making it well-suited for cluttered warehouse scenes where dense semantic segmentation may be infeasible.

In addition to this, Gonzalez et al.~\cite{Dimitrios} propose a dual function algorithm that combines human localization and free space segmentation. Using sparse point cloud representations and region-growing techniques, their system reliably identifies navigable ground and potential human obstacles with minimal computational overhead. Designed for deployment on mobile robots in warehouse facilities, their method emphasizes safety and efficiency in real-world industrial environments.


\subsection{Contributions} \label{sec_contrib}
In the state of the art, various approaches have been presented for LiDAR semantic segmentation.
However, all of these methods are developed for road traffic scenarios. Transferring concepts from the traffic domain to industrial warehouse environments remains a challenge and is largely unexplored.
In this context, the main contributions of this paper are:
\begin{itemize}
\item A method for 3D semantic segmentation using data from a dual-LiDAR setup mounted on a counterbalance forklift.
\item An empirical study of LiDAR-based semantic segmentation in real-world industrial warehouse environments.
\end{itemize}

\subsection{Outline} \label{sec_outline}
The remainder of this article is structured as follows:
In \autoref{sec:dataset}, the dataset and the data collection process are described. Followed by our method for semantic perception in \autoref{sec_method}. Our results are reported in \autoref{sec:expirement}, followed by a conclusion and short outlook in \autoref{sec:conclution}.

\section{\large Dataset}
\label{sec:dataset}
\begin{figure}[!h] 
    \centering
    \includegraphics[width=\columnwidth]{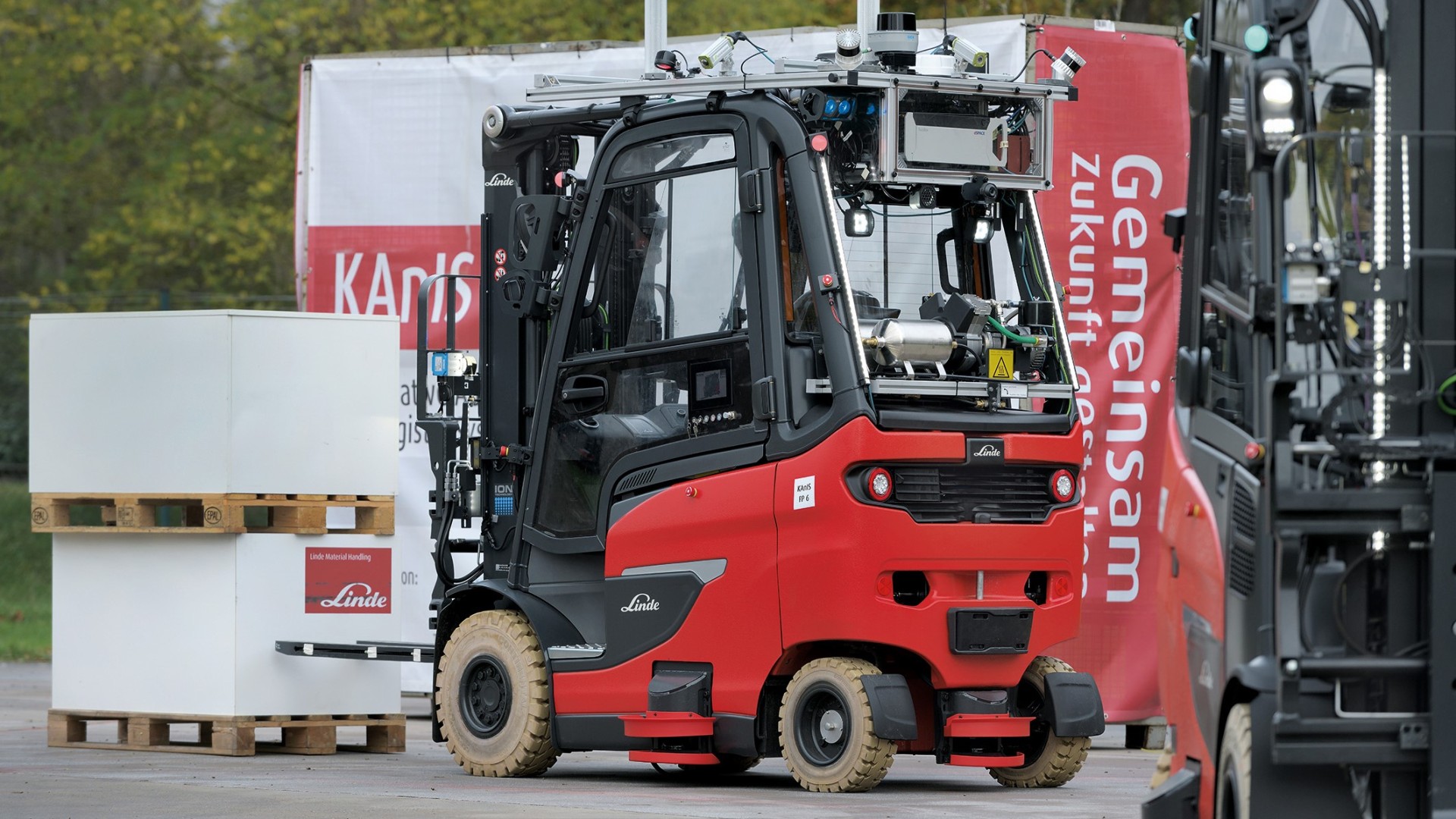}
    \caption{Prototype forklift in action. © Linde Material Handling}
    \label{fig:SensorSetup}
\end{figure}
Figure \ref{fig:SensorSetup} shows the recording system used in action. For our data collection, we utilize the two Ouster OS0-128 LiDAR sensors mounted on the roof \cite{ouster_os0}. The sensors are synchronized using the precision time protocol (PTP) with GPS time as reference. 
We recorded our data at the Linde Material Handling factory site during normal work operations.
For the annotation process, we utilize \textit{PointLabeler} by J. Behley et al. \cite{behley2019iccv}, which supports the grouping and annotation of multiple LiDAR scans through ego motion compensation. To estimate ego-motion, we employ the Ouster SDK SLAM module, which incorporates a high-level implementation of \textit{KISS-ICP} \cite{kiss-icp}. 
\begin{table}[!h]
\centering
\caption{Dataset Scenes}
    \label{tab:scenes}
\begin{tabular}{l|llll}
Sequence & Frames & Scans & Points & Split \\ 
\midrule
0000     & 665  & 1330 & 348.6 M & Test \\ 
0001     & 667  & 1334 & 349.7 M & Train \\ 
0002     & 723  & 1446 & 379.1 M & Train \\ 
0003     & 1783  & 3566 & 934.8 M & Train \\ 
\end{tabular}
\end{table}
We collected a total of 7676 LiDAR scans in four sequences with a total of 2012.2 million annotated 3D points. Sequence 0000 is reserved for testing, while the remaining sequences are used for training (see \autoref{tab:scenes}). Due to the nature of the factory site, we recorded a large number of forklifts and material handling utilities from various perspectives. We defined our classes based on their relevance to material handling resulting in the following semantic classes:
\begin{table}[!h]
\centering
\caption{Class Definitions}
    \label{tab:class_def}
\begin{tabular}{l|l}
Class & Definition  \\ 
\midrule
\textit{car} & Vehicle designed for road traffic. \\
\textit{forklift} & Forklift used for material handling. \\
\textit{person} & Human pedestrian. \\
\textit{object} & Movable object or temporary obstacle. \\
\textit{driveable ground} & Surface suitable for AGV navigation.\\
\textit{other ground} & Surface not suitable for AGV navigation.\\
\textit{lane marking} & Painted or marked lane indicator on the ground. \\
\textit{vegetation} & Natural plant life including grass, shrubs, and trees. \\
\textit{building} & Static man-made structure.\\
\end{tabular}
\end{table}

\begin{figure}[!h] 
    \centering
    \includegraphics[width=\columnwidth]{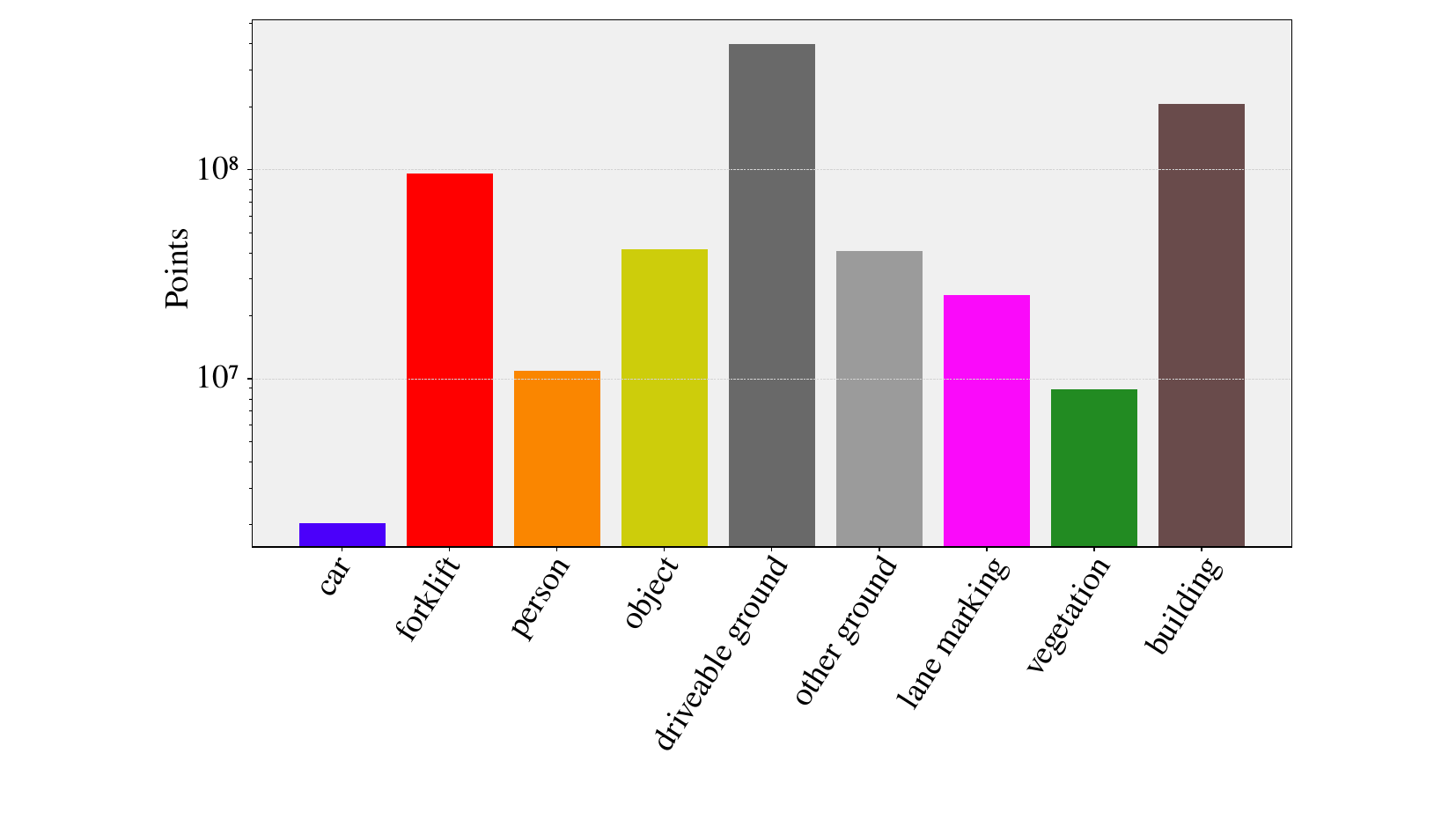}
    \caption{Distributions of semantic classes in SemanticTHAB.}
    \label{fig:SemanticTHAB_statistics}
\end{figure}

\begin{figure}[!h] 
    \centering
    \includegraphics[width=\columnwidth]{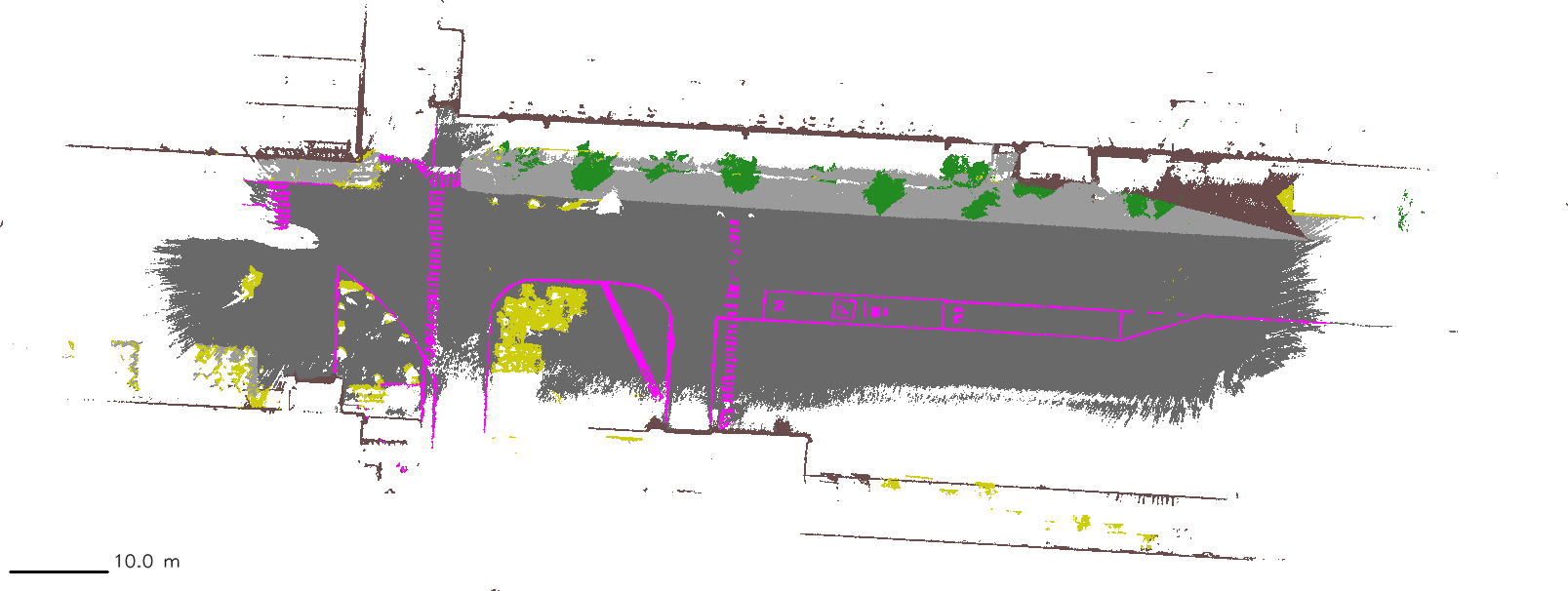}
    \caption{Reference Map of Scene 0000.}
    \label{fig:referencemap}
\end{figure}

The distribution of the classes in our data set can be seen in \autoref{fig:SemanticTHAB_statistics}. Note that the figure is using logarithmic scale.
We created a reference map of scene 0000 using VDBfusion \cite{vizzo2022sensors} as shown in \autoref{fig:referencemap}. We remove all dynamic objects for better visualization.

\section{\large Method}\label{sec_method}

The following section contains an explanation of the methods used in our perception pipeline.
Our method consists of two main steps. First, our data preparation process is detailed in \autoref{sec:Prepossessing}, which includes structuring the LiDAR point clouds into an image-like format, transforming the scans into a defined forklift coordinate system, and computing surface normals. Second, we present our core semantic segmentation method in \autoref{sec:semantic_segmentation}.

\subsection{Preprocessing} \label{sec:Prepossessing}
\begin{figure}[!h] 
    \centering
    \includegraphics[width=0.5\columnwidth]{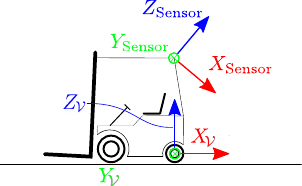}
    \caption{The forklift vehicle frame (following ISO 8855 conventions) and the LiDAR sensor coordinate frame.}
    \label{fig:ISO8855}
\end{figure}
Our method uses a staggered, image-like LiDAR representation based on spherical projection, following \cite{2023.reichert, reichert2025realtimesemanticsegmentation}. Each 3D point $\vec{p} = [x, y, z]^T$ is transformed into spherical coordinates $\vec{x} = [\phi, \theta, r]^T$, where $\phi$ is the azimuth, $\theta$ the inclination, and $r$ the range. These are projected into image coordinates $(u, v)$ using:

\begin{equation}
\begin{bmatrix}
u\\
v\\
1
\end{bmatrix}
=
\begin{bmatrix}
\frac{1}{\Delta \phi} & 0 & c_\phi\\
0 & \frac{1}{\Delta \theta} & c_\theta\\
0 & 0 & 1
\end{bmatrix}
\cdot
\begin{bmatrix}
\phi\\
\theta\\
1
\end{bmatrix}
\end{equation}

\begin{figure*}[!h]
    \centering
    \begin{minipage}{\textwidth}
        \centering
        \includegraphics[width=\textwidth]{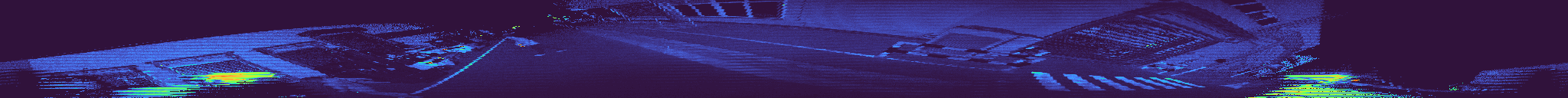}
        \label{fig:kitti}
    \end{minipage}
    \begin{minipage}{\textwidth}
        \centering
        \includegraphics[width=\textwidth]{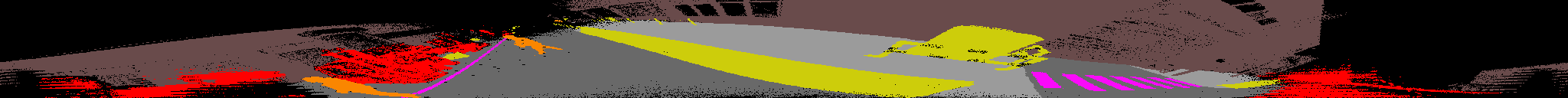}
        \label{fig:kitti}
    \end{minipage}
    \begin{minipage}{\textwidth}
        \centering
        \includegraphics[width=\textwidth]{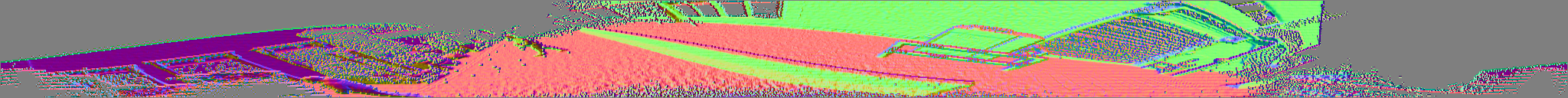}
        \label{fig:surface_normals}
    \end{minipage}
     \vspace{-4mm}
    \caption{Spherical projections of reflectivity measurements (top), semantic annotations (middle), and surface normals (bottom)}
    \label{fig:reflectivity} 
\end{figure*}

This generates spherical images $I_{x,y,z}(u,v)$, reflectivity maps $I_{\textit{ref}}(u,v)$ (see \autoref{fig:reflectivity}), and range images $I_r(u,v)$ via Euclidean norm. In addition, we can transform the semantic annotation $I_{\textit{sem}}(u,v)$ into a spherical image.

The points in the point cloud and their corresponding pixels are transformed to the vehicle coordinate system $\mathcal{V}$ defined in ISO8855 \cite{ISO8855} using a transformation matrix:
\begin{equation}
    {}^{\mathcal{V}}I_{x,y,z}(u,v) = \begin{bmatrix}
        R & t
    \end{bmatrix}  I_{x,y,z}(u,v)
    \label{eq:transform}
\end{equation}

This transformation is crucial because it assumes that the ground of the immediate surroundings is parallel to the vehicle's $xy$-plane. The ISO8855 coordinate system defines the center rear wheel axis as the origin and the $x$ axis is aligned with the movement direction as shown in \autoref{fig:ISO8855}. Using a joint coordinate system is essential for the fusion of both LiDAR sensors in our perception approach.

Surface normals provide a local geometric context for segmentation and obstacle detection. Following \cite{surface_normals, reichert2025realtimesemanticsegmentation}, we estimate normals using finite differences. For each pixel $P_c = {}^{\mathcal{V}}I_{x,y,z}(u,v)$ and its neighbors $P_b = {}^{\mathcal{V}}I_{x,y,z}(u+1,v)$, $P_a = {}^{\mathcal{V}}I_{x,y,z}(u,v+1)$, the normal is computed as:

\begin{equation}
{}^{\mathcal{V}}I_{n_x,n_y,n_z}(u,v) = \frac{(P_b - P_c) \times (P_a - P_c)}{\|(P_b - P_c) \times (P_a - P_c)\|_2}
\end{equation}

This produces a dense surface normal map $I_{n_x,n_y,n_z}(u,v)$ for downstream geometric processing. Examples of surface normals can be seen in \autoref{fig:reflectivity}.
\subsection{Semantic Segmentation} \label{sec:semantic_segmentation}
\begin{figure}[!h] 
    \centering
    \includegraphics[width=\columnwidth]{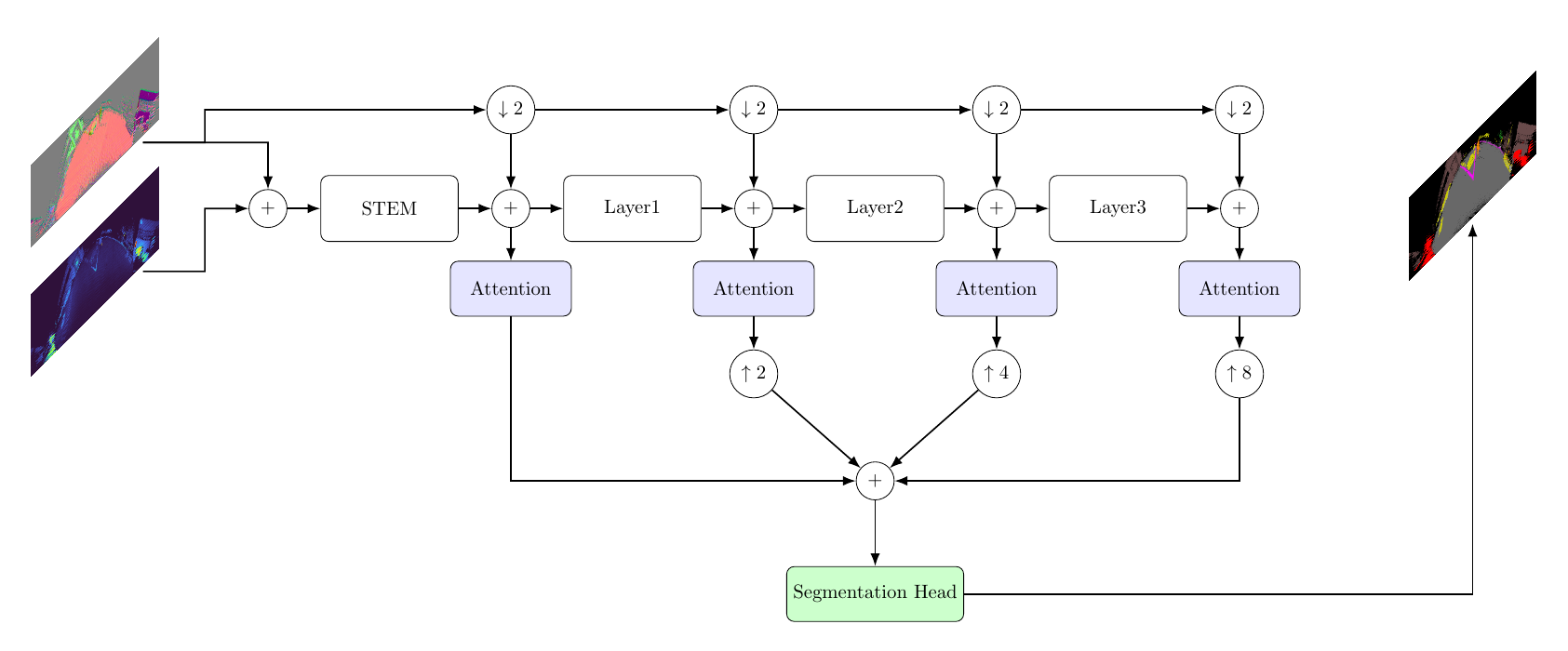}
    \caption{\textit{\textbf{Model Architecture}}: white blocks indicate the backbone path; $\downarrow$ denotes nearest-neighbor downsampling; $+$ represents feature concatenation; $\uparrow$ corresponds to deconvolution-based upsampling; the green block indicates the segmentation head.}
    \label{fig:network_arch}
    \vspace{-5mm} 
\end{figure}

For our semantic segmentation module, we use CNNs as it is evident from \autoref{sec_sota} that CNNs are well suited for real-time processing of spinning LiDAR scans.
As input, we utilize the spherical projection of the point cloud expressed in the forklift coordinate system, denoted ${}^{\mathcal{V}}I_{x,y,z}(u,v)$, the reflectivity measurements represented as the spherical image $I_{\textit{ref}}(u,v)$, and the surface normals ${}^{\mathcal{V}}I_{n_x,n_y,n_z}(u,v)$. We assume that reflectivity measures $I_{\textit{ref}}(u,v)$ are essential for detecting classes such as \textit{lane markings}, which are visible exclusively in reflectivity data. Furthermore, we assume that ${}^{\mathcal{V}}I_{n_x,n_y,n_z}(u,v)$ enhances the network's ability to identify structures by leveraging their orientation and homogeneity. We follow the insights of \cite{Alonso20203DMiniNetLA} and include ${}^{\mathcal{V}}I_{x,y,z}(u,v)$ and ${}^{\mathcal{V}}I_{n_x,n_y,n_z}(u,v)$ downsized to the respective resolution at any stage where the spatial resolution of the latent feature map changes to improve 3D consistency. For the neck, we first employ multiplicative self-attention \cite{NIPS2017_3f5ee243} by using multi-scale features extracted from the backbone.
The attention-weighted features are then upsampled using deconvolution layers to the shape  \( H/2 \times W/2 \) (i. e. \( 64 \times 1024 \)) and merged by a concatenation operation to build a feature pyramid network (FPN) \cite{8099589}. FPN enhances the ability of network to detect objects of vastly different sizes by combining low-resolution, semantically strong features with high-resolution, semantically weak features. For the segmentation head, we use a deconvolution layer to reshape the output to \( H \times W \) (i. e. \( 128 \times 2048 \)), followed by two convolutional layers for anti-aliasing. 
Our network architecture is visualized in \autoref{fig:network_arch}. As backbones, we support three variants from the \textit{ResNet} family (\textit{ResNet18}, \textit{ResNet34}, and \textit{ResNet50})\cite{ResNet}, as well as three variants from the \textit{ShuffleNet} family \cite{ShuffleNet} and three backbones from the \textit{EfficientNet} family \cite{EfficientNetV2}.

\subsection{Implementation Details}
\subsubsection{Training Objectives}
For the loss functions, we use a weighted sum of the cross-entropy loss \( \mathcal{L}_{CE} \) and the Tversky loss \( \mathcal{L}_{\text{Tversky}} \) \cite{Hashemi2018TverskyAA}. The Tversky loss allows for adjusting the balance between false positives and false negatives, which is particularly useful for detecting smaller objects. During training, we treat each of the two sensors separately by randomly mixing data from both sensors into the training batches.
\subsubsection{Training Curriculum}
Our models were trained with a batch size of 8 and a learning rate of 0.001, using the ADAM optimizer \cite{Kingma2014AdamAM}. To ensure stable training and convergence, we incorporated a learning rate scheduler. All models were initially pre-trained on the SemanticKITTI dataset for 50 epochs, followed by fine-tuning on our dataset for 30 epochs. We trained our model on all sequences except 0000, which we use for testing.
\subsubsection{Hardware Setup}
We use a single Nvidia RTX 3090 for training and testing. If not explicitly stated differently, all run times are measured using a RTX 3090.
\subsubsection{Inference}
During inference, we operate our semantic segmentation model with a batch size of two, i. e. one for each LiDAR sensor. 

\section{\large Results}
\label{sec:expirement}
The standard metric for evaluating semantic segmentation is the mean
Intersection over Union (mIoU). For each class $c$, it compares the prediction region with the ground truth region, resulting in
\begin{equation}
    IoU_c = \frac{\sum_{u,v} \mathds{1}(I_{\textit{sem}}^{Pred}(u,v)= c \wedge I_\textit{sem}^{GT}(u,v)=c)}{\sum_{u,v} \mathds{1}(I_{\textit{sem}}^{Pred}(u,v)= c \vee I_\textit{sem}^{GT}(u,v)=c)}
\end{equation}
with $I_{\textit{sem}}^{Pred}(u,v)$ being the predicted label, $I_\textit{sem}^{GT}(u,v)$ the ground truth and $\mathds{1}$ the indicator function. The IoU is
averaged over the entire test set and all
classes to obtain the mIoU.

We conducted experiments using different backbone architectures to evaluate semantic segmentation accuracy and real-time performance. Two processing setups are considered: single LiDAR scan inference with a batch size of 1, and dual-sensor inference with a batch size of 2.

\begin{figure}[!h] 
    
    \centering
    \includegraphics[width=\columnwidth]{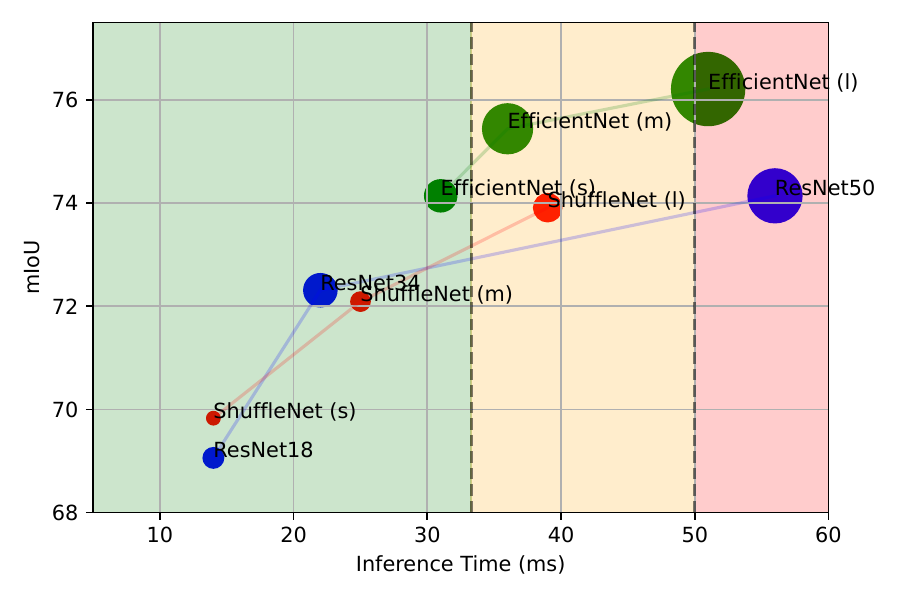}
    \caption{\textit{\textbf{Model Comparison}}: mIoU and inference time at single sensor inference. The size of the points sketch the number of parameters. The different zones show the real-time constraints.}
    \label{fig:inference_time}
\end{figure}

Models are assessed under a real-time constraint of 33.3 ms for dual inference. As shown in \autoref{tab:miouresults_median} and \autoref{fig:inference_time}, \textit{EfficientNet (s)} achieves the best trade-off, reaching 74.14\% mIoU, 31 ms latency, and 22.69M parameters. While \textit{EfficientNet (l)} offers higher accuracy (76.21\% mIoU), its latency exceeds the real-time threshold. Models based on \textit{ResNet} and \textit{ShuffleNet} generally perform worse in speed or accuracy.

Application-specific priorities also influence the selection of the model. In our case, accurate segmentation of the ground, person, and forklift classes is crucial. Lightweight models such as \textit{ResNet18} and \textit{ShuffleNet (s)} already perform well in these categories, achieving forklift IoUs between 85.51\% and 88.93\% and driveable ground IoUs of 92.32\% and 91.03\%, respectively. Higher-capacity models like those in the \textit{EfficientNet} family significantly improve person segmentation, reaching up to 80.04\% IoU.

\begin{figure*}[!htb]
    \centering
    \begin{minipage}[b]{0.3\textwidth}
        \centering
        \includegraphics[width=\textwidth]{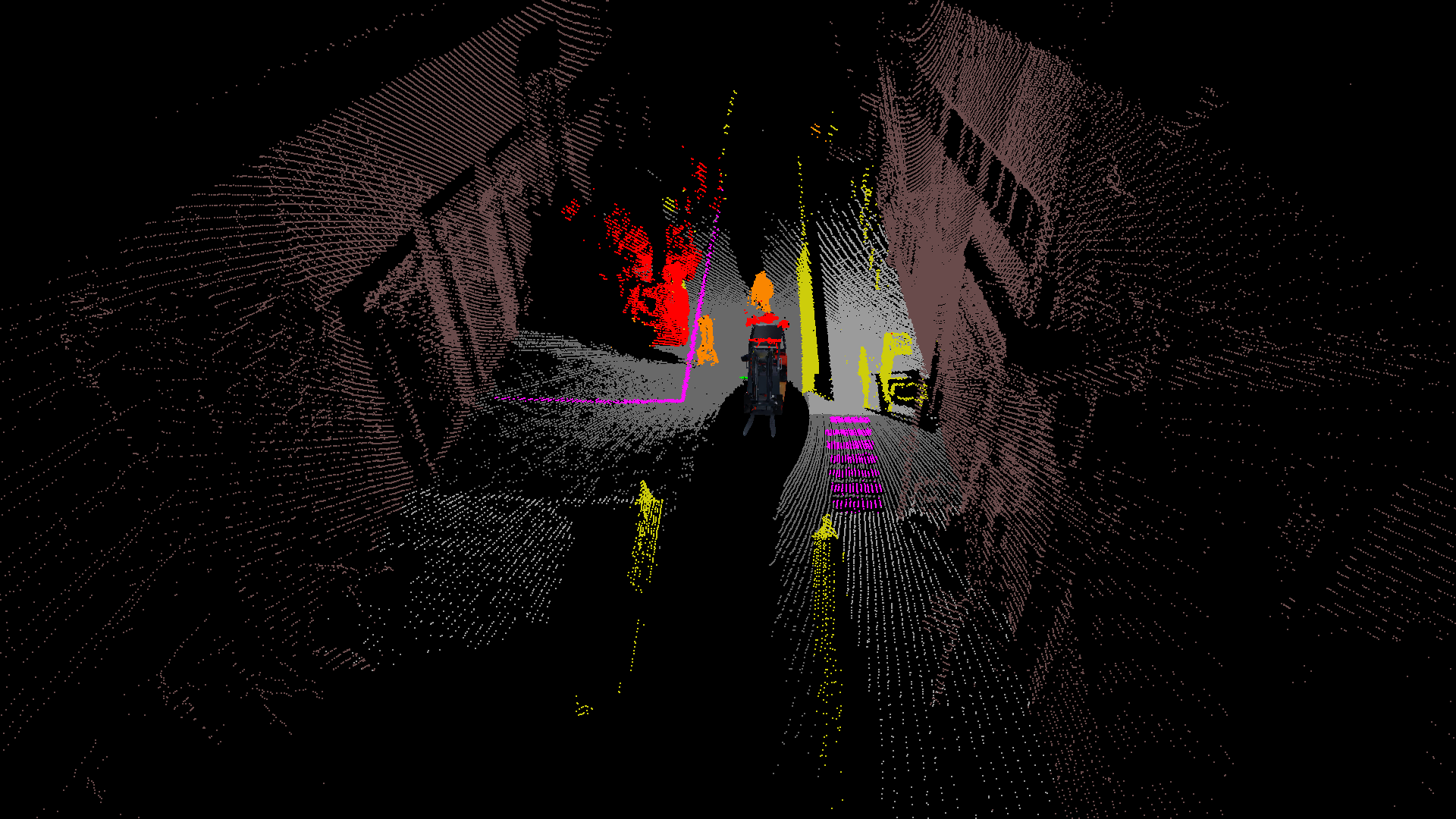}

    \end{minipage}
    \hspace{0.02\textwidth} 
    \begin{minipage}[b]{0.3\textwidth}
        \centering
        \includegraphics[width=\textwidth]{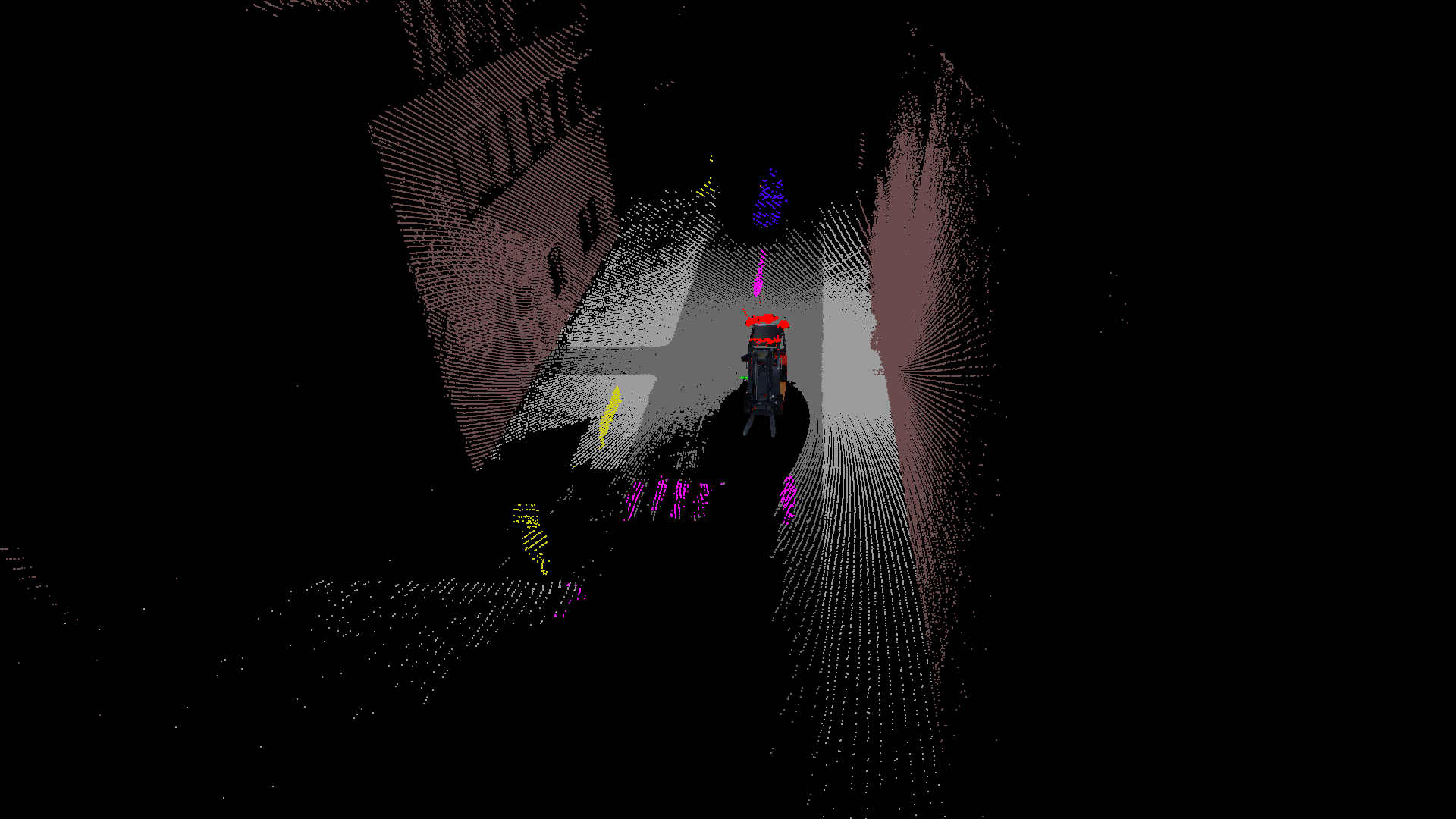}
    \end{minipage}
    \hspace{0.02\textwidth} 
    \begin{minipage}[b]{0.3\textwidth}
        \centering
        \includegraphics[width=\textwidth]{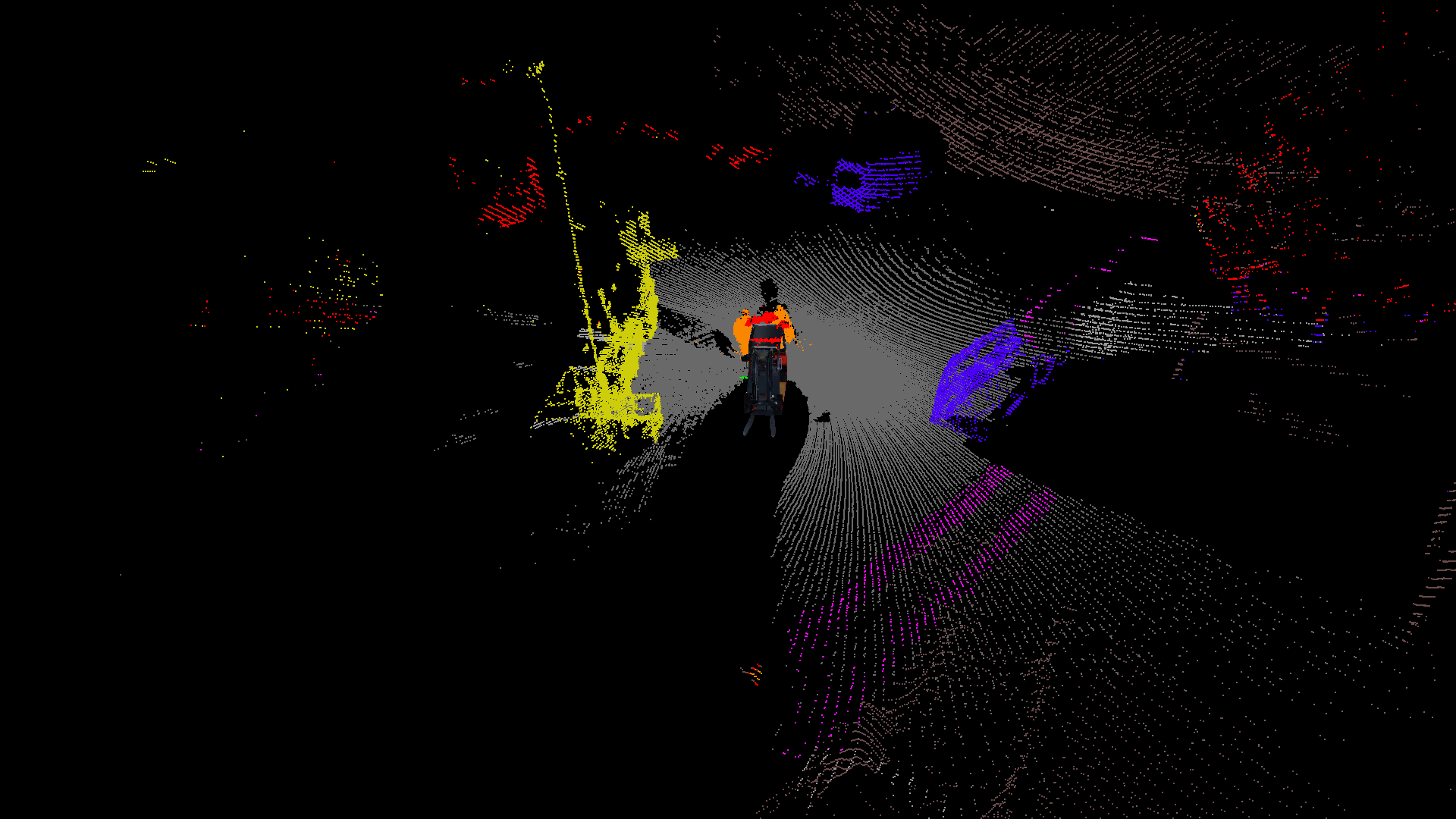}
    \end{minipage}

\caption{\textit{\textbf{Qualitative Results}}: Semantic segmentation of point clouds. Lane markings (purple) are clearly distinguishable due to the use of reflectivity measurements. Drivable and non-drivable ground are well separated, particularly in areas with elevation differences. Persons (red), cars (blue), and forklifts (orange) are segmented with sufficient accuracy to be easily recognizable.}

\label{fig:quant_results}
\vspace{-5mm} 
\end{figure*}

The qualitative results in \autoref{fig:quant_results} show the effectiveness of the model in real forklift operations within outdoor warehouse environments. The strong distinction between drivable and non-drivable ground is likely due to the integration of surface normal features, which capture terrain geometry effectively. Lane markings are segmented with high precision, aided by LiDAR reflectivity, which enhances visibility under varying environmental conditions. Dynamic objects such as people, forklifts, and vehicles are consistently detected, supporting robust obstacle awareness essential for collision avoidance and emergency response in complex scenarios.

\begin{table*}[t]
\centering
\caption{\textit{\textbf{Quantitative Results}}: Comparison of different backbones in terms of model size (Parameters), inference time (single and dual), per-class IoU for selected classes, and overall mean IoU (mIoU)}
\label{tab:miouresults_median}
\resizebox{\textwidth}{!}{
\begin{tabular}{|c|l|ll|cccccc|c|}
\toprule
Backbone & Parameters &  \multicolumn{2}{c|}{Inference Time} & forklift & person & object & driveable & lane & building & mIoU \\
 &  & single & dual &  &  &  & ground & marking &  &  \\
\midrule
ResNet18 & 9.03M & \textbf{8ms} & \textbf{14ms} & 85.51 & 62.50 & 56.65 & 92.32 & 26.87 & 91.67 & 69.06\\
ResNet34 & 24.27M & 12ms & 22ms  & 88.93 & 73.01 & 59.77 & 92.47 & 39.91 & 91.58 &  72.31 \\
ResNet50 & 65.15M & 30ms & 56ms  & 93.65 & 79.07 & 60.96 & 92.91 & 40.61 & 91.98 &  74.14 \\
ShuffleNet (s) & \textbf{3.73M} & 12ms & \textbf{14ms} & 87.32 & 61.21 & 57.32 & 91.03 & 32.31 & 91.13 &  69.83 \\
ShuffleNet (m) & 7.98M & 15ms & 25ms  & 91.14 & 75.46 & 56.7 & 92.92 & 32.50 & 90.90 &  72.09 \\
ShuffleNet (l) & 17.41M & 21ms & 39ms  & \textbf{96.72} & 78.3 & 61.66 & 92.79 & 40.90 & 92.62 &  73.91\\
EfficientNet (s) & 22.69M & 15ms & 31ms  & 95.02 & 79.48 & 59.44 & 93.63 & 39.39 & 91.89 &  74.14 \\
EfficientNet (m) & 55.60M & 19ms & 36ms  & 96.21 & 79.89 & 61.9 & \textbf{93.92} & 41.07 & \textbf{92.79} & 75.44 \\
EfficientNet (l) & 120.31M & 27ms & 51ms  & 95.37 & \textbf{80.04} & \textbf{62.23} & 93.01 & \textbf{42.08} & 92.53 &  \textbf{76.21} \\

\bottomrule
\end{tabular}
}
\end{table*}

\section{\large Conclusions and Future Work}
\label{sec:conclution}

In this work, we presented a novel semantic segmentation framework that leverages a dual high-resolution LiDAR setup specifically designed for autonomous forklifts operating in complex outdoor warehouse environments. Our approach demonstrated the ability to accurately segment safety-critical classes such as pedestrians, vehicles, and forklifts, as well as environmental characteristics crucial for navigation, including drivable ground and lane markings. By combining forward-facing and downward-angled LiDAR sensors, our system achieved enhanced spatial coverage and robustness under real-world operational conditions. The lightweight segmentation method enables real-time inference, meeting the stringent timing requirements of autonomous forklift control loops.

Future work will focus on extending the current framework to incorporate dynamic scene understanding through temporal fusion, improving the tracking of moving obstacles, and handling occlusions more effectively. Additionally, integrating multi-modal sensor data such as camera imagery and radar could further enhance semantic perception under adverse weather or lighting conditions. Finally, deploying the system in a closed-loop autonomous navigation pipeline will provide comprehensive validation of its impact on task efficiency and safety in operational environments.



{\small
\bibliographystyle{ieee_fullname}
\bibliography{egbib}
}

\end{document}